\documentclass{article}

\usepackage{microtype}
\usepackage{graphicx}
\usepackage{subfigure}
\usepackage{booktabs} 

\usepackage{hyperref}

\usepackage[accepted]{icml2024}

\usepackage{amsmath}
\usepackage{amssymb}
\usepackage{mathtools}
\usepackage{amsthm}

\usepackage[capitalize,noabbrev]{cleveref}

\theoremstyle{plain}

\theoremstyle{definition}

\theoremstyle{remark}

\usepackage[textsize=tiny]{todonotes}

\icmltitlerunning{Nonlinear model reduction with CoLoRA}

\newcommand{\mathdefault}[1][]{}
\usepackage{amsfonts,amsmath,amssymb,mathtools}      
\usepackage{nicefrac}      
\usepackage{microtype}     
\usepackage{graphicx}
\usepackage{parskip}
\usepackage{algorithm}
\usepackage{algorithmic}
\usepackage{xcolor}
\usepackage{pgf}
\usepackage{lipsum}
\usepackage{anyfontsize}
\usepackage{tikz}
\usepackage{layouts}
\usepackage{array}
\usepackage{kantlipsum}
\usepackage[mode=buildnew,subpreambles=true]{standalone}
\usepackage{xspace}
\usepackage{siunitx}
\usepackage{import}
\usepackage{enumitem}

\newcommand{\A}{\boldsymbol A} 
\newcommand{\B}{\boldsymbol B} 
\newcommand{\W}{\boldsymbol W} 
 
\newcommand{\rank}{r} 
\newcommand{\colora}{\mathcal{C}}

\newcommand{\clayers}{L}
\newcommand{\dimReduced}{q}
\newcommand{\dimParamsOff}{p}
\newcommand{\dimHyper}{\dimReduced^{\prime}}

\newcommand{\fom}{u_{\text{F}}}
\newcommand{\rom}{{\hat{u}}}

\renewcommand{\t}{t}
\newcommand{\Tend}{T}
\newcommand{\m}{ \boldsymbol {\mu}}
\newcommand{\x}{ \boldsymbol{x}}
\newcommand{\tSpace}{\mathcal T}
\newcommand{\mSpace}{\mathcal D}
\newcommand{\xSpace}{\Omega}
\newcommand{\nt}{K}
\newcommand{\nm}{m}
\newcommand{\nx}{n_x}

\newcommand{\dimX}{d}

\newcommand{\bfx}{\boldsymbol{x}}
\newcommand{\bfb}{\boldsymbol{b}}

\newcommand{\bfmu}{\boldsymbol{\mu}}

\DeclareMathOperator{\diag}{diag}
\newcommand{\R}{\mathbb{R}}

\newcommand{\paramsOff}{ \boldsymbol {\theta}}
\newcommand{\paramsOn}{\boldsymbol {\phi}}
\newcommand{\paramsHyper}{\boldsymbol {\psi}}

\newcommand{\xSamples}{\mathcal{X}}
\newcommand{\tSamples}{\mathcal{T}}

\newcolumntype{L}{>{$}l<{$}} 

\begin{document}

\twocolumn[
\icmltitle{CoLoRA: Continuous Low-Rank Adaptation for Reduced Implicit Neural Modeling of Parameterized Partial Differential Equations}



\icmlsetsymbol{equal}{*}

\begin{icmlauthorlist}
\icmlauthor{Jules Berman}{yyy}
\icmlauthor{Benjamin Peherstorfer}{yyy}
\end{icmlauthorlist}

\icmlaffiliation{yyy}{Courant Institute of Mathematical Sciences, New York University, New York, NY 10012, USA}

\icmlcorrespondingauthor{Jules Berman}{jmb1174@nyu.edu}

\icmlkeywords{partial differential equations, reduced modeling, nonlinear model reduction, implicit neural representations, pre-training, fine-tuning}

\vskip 0.3in
]





\printAffiliationsAndNotice{} 

\begin{abstract}
This work introduces reduced models based on Continuous Low Rank Adaptation (CoLoRA) that pre-train neural networks for a given partial differential equation and then continuously adapt low-rank weights in time to rapidly predict the evolution of solution fields at new physics parameters and new initial conditions. The adaptation can be either purely data-driven or via an equation-driven variational approach that provides Galerkin-optimal approximations. Because CoLoRA approximates solution fields locally in time, the rank of the weights can be kept small, which means that only few training trajectories are required offline so that CoLoRA is well suited for data-scarce regimes. Predictions with CoLoRA are orders of magnitude faster than with classical methods and their accuracy and parameter efficiency is higher compared to other neural network approaches.
\end{abstract}

\section{Introduction}\label{sec:Intro}
Many phenomena of interest in science and engineering depend on physics parameters $\m$ that influence the temporal and spatial evolution of the system such as the Reynolds number in fluid mechanics and conductivity coefficients in heat transfer. %
Rapidly simulating physical phenomena for a large sample $M \gg 1$ of physics parameters $\m_1, \dots, \m_M$ is paramount in science and engineering, e.g., for finding optimal designs, inverse problems, data assimilation, uncertainty quantification, and control.
Numerically solving the underlying parameterized partial differential equations (PDEs) with standard numerical methods \cite{Hughes2012,LeVeque_2002} for large numbers of different physics parameters is prohibitively expensive in many applications and thus one often resorts to reduced modeling. %

\begin{figure*}[t]
\centering
\includegraphics[width=0.8\textwidth]{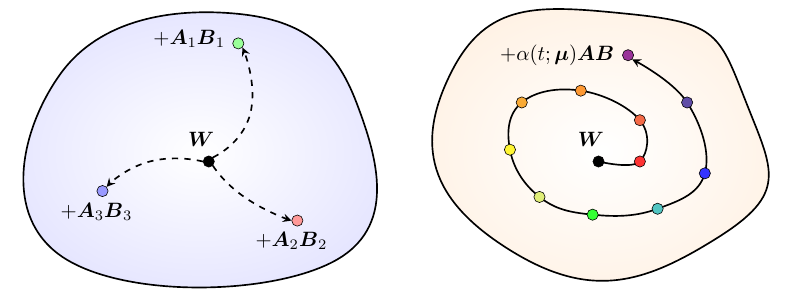}

\caption{LoRA fine-tunes networks to downstream tasks by adapting low-rank matrices $\A \B$. Our CoLoRA introduces a scaling $\alpha(t, \m)$ on the low-rank matrix $\A \B$ to adapt networks continuously to predict PDE solution trajectories.} 
\label{fig:CoLoRAOverview}
\end{figure*}

\paragraph{The Kolmogorov barrier}
Reduced models exploit structure in PDE problems to more efficiently approximate solution fields. The conventional structure that is leveraged is low rankness in the sense of the classical principal component analysis. Reduced models based on such low-rank approximations can achieve exponentially fast error decays $\mathrm e^{-c n}$ with the rank $n$ for a wide range of (mostly nicely behaved elliptic) problems \cite{MADAY2002289,10.1093/imanum/dru066}.
However, linear low-rank approximations are affected by the so-called Kolmogorov barrier, which states that there are classes of PDEs for which linear approximations have an error decay rate of at best $1/\sqrt{n}$ \cite{Ohlberger16,GREIF2019216}. Examples of PDE classes that are affected by the Kolmogorov barrier are often describing transport-dominated phenomena such as strongly advecting flows and wave-like behavior; see  \cite{P22AMS} for a survey.

\begin{figure*}[t]
\centering
\import{plots}{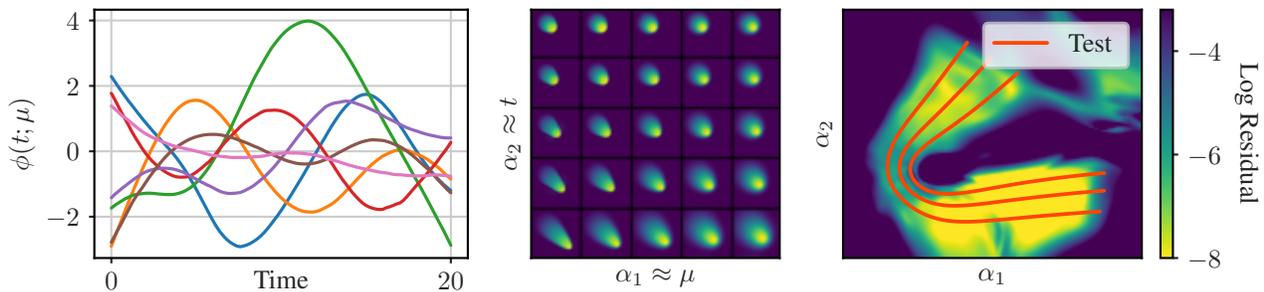}
\caption{\textbf{Left:} Shows that CoLoRA's latent states  $\paramsOn(t; \m)$ adapt smoothly over time (RDE example). 
\textbf{Middle:} Training a CoLoRA model with a $\dimReduced = 2$-dimensional latent state on the Burgers' example gives the first latent component corresponding to translation in time and the second one to the viscosity $\mu$. 
\textbf{Right:}  CoLoRA learns a continuous region of low PDE residual along which the latent trajectories evolve (Vlasov example); see Appendix~\ref{appx:num}.} 
\label{fig:motivation}
\end{figure*}

\paragraph{Our contribution: nonlinear reduced modeling with continuous low-rank adaptation (CoLoRA)}
In this work, we build on LoRA (Low Rank Adaptation) for developing parameterizations for reduced models \cite{hu_lora_2021}.
LoRA has been developed for fine-tuning large language models and leverages the observation that fine-tuning objectives can be efficiently optimized on low-dimensional parameter spaces \cite{li2018measuring,aghajanyan-etal-2021-intrinsic,hu_lora_2021}. This has been observed not only for large language models but also when approximating solution fields of physical phenomena \cite{10.1093/imanum/drx052,https://doi.org/10.1002/gamm.201310004,berman_randomized_2023}.
We build on the pre-training/fine-tuning paradigm of LoRA but modify it to Continuous LoRA (CoLoRA) that reflects our PDE setting by allowing continuous-in-time adaptation (``fine-tuning'') of parts of the low-rank components as the solution fields of the PDEs evolve; see Figure~\ref{fig:CoLoRAOverview}. This inductive bias is in agreement with how typically physical phenomena evolve over time, namely smoothly and along a latent low-dimensional structure.
By composing multiple CoLoRA layers in a deep network, we obtain a nonlinear parameterization that can circumvent the Kolmogorov barrier while having a pre-training/fine-tuning decomposition with an inductive bias that reflects the special meaning of the time variable $t$ in PDE problems; see Figure~\ref{fig:motivation}. %

\paragraph{The time variable in CoLoRA models}
Time enters CoLoRA models via the low-rank network weights that are adapted in the online (``fine-tuning'') phase of reduced modeling, rather than as input to the network as the spatial coordinates.
Treating time separately from inputs such as the spatial coordinates aligns well with approximating PDE solution fields sequentially in time, rather than globally over the whole time-space domain. The sequential-in-time approximation paradigm of CoLoRA allows keeping the number of offline and online parameters low, because the solution field is approximated only locally in time.
Requiring a low number of online and offline parameters indicates that the inductive bias induced by separating out the time variable from the spatial coordinate is in agreement with many physics problems.
Furthermore, the CoLoRA architecture can be trained on low numbers of training trajectories compared to, e.g., operator learning \cite{li_fourier_2021,Lu2021}, which is important because standard numerical simulations can be expensive and thus only a limited number of training trajectories are available in many cases. 
Having network weights that depend on time allows combining CoLoRA with equation-driven variational approaches \cite{lasser_lubich_2020,Du_2021,doi:10.1137/21M1415972,bruna_neural_2024,WEN2024134129} to obtain reduced solutions that are Galerkin-optimal, which opens the door to analyses, error bounds, and goes far beyond purely data-driven forecasting.

\paragraph{Literature review}
\label{sec:litreview}
There are purely data-driven surrogate modeling methods such as operator learning \cite{li_fourier_2021,Lu2021,boulle_mathematical_2023} that can require large amounts of data because they aim to learn a generic operator map over the full model space.
Model reduction  \cite{antoulas2005approximation,RozzaPateraSurvey,benner2015survey,interpbook,doi:10.1146/annurev-fluid-121021-025220} considers a more structured problem via the physics parameter $\m$, for which nonlinear model reduction methods based on autoencoders have been presented in \cite{carlburg_lee_2020,Lee_Carlberg_2021,KIM2022110841,https://doi.org/10.48550/arxiv.2203.00360}; however, they can require going back to the high-fidelity numerical model to drive the dynamics, which can be expensive. Alternative approaches learn the low-dimensional latent dynamics \cite{Fulton:LSD:2018,doi:10.1098/rspa.2021.0162,wan2023evolve}. Additional literature of nonlinear model reduction is reviewed in Appendix~\ref{appx:MORLiterature}. 

There is a range of methods based on implicit neural representations that updates the network parameters either based on the equation \cite{chen2023implicit,chen_crom_2023} or via hyper-networks \cite{brunton_implcitflow,yin_continuous_2023}; these methods are closest to CoLoRA, except that we will show that CoLoRA's low-rank adaptation achieves lower errors with lower parameter counts in our examples. %
There are pre-training/fine-tuning techniques for global-in-time methods such as physics-informed neural networks \cite{RAISSI2019686} that are purely data-driven once trained. Some of these approaches use hyper-networks \cite{belbute-peres2021hyperpinn,cho2023hypernetworkbased} and other meta-learning such as \citet{pmlr-v70-finn17a}. But typically these approaches are over time-space domains and thus do not make a special treatment of time or the PDE dynamics, which is a key feature of CoLoRA. Adaptive low-rank approximations have been used in scientific computing for a long time \cite{doi:10.1137/050639703,SAPSIS20092347,Peherstorfer15aDEIM,doi:10.1137/18M1218686}; however, they use one layer only and adapt the low-rank matrices directly with time. We have a more restricted adaptation in time that has fewer parameters, which is sufficient due to the nonlinear composition of multiple layers. Other low-rank approximations have been widely used in the context of deep networks \cite{6638949,6853583,7472630,khodak2021initialization,NEURIPS2022_7e98b00e} but not in PDE settings. %

\paragraph{Summary} CoLoRA leverages that PDE dynamics are typically continuous in time while evolving on low-dimensional manifolds. With CoLoRA, we achieve nonlinear parameterizations that circumvent the Kolmogovorv barrier of transport-dominated problems and can provide predictions purely data-driven or in a variational sense using the PDEs. Our numerical experiments show that CoLoRA requires a low number of training trajectories, achieves orders of magnitude speedups compared to classical methods, and outperforms the existing state-of-the-art neural-network-based model reduction methods in parameter count and accuracy on a wide variety of PDE problems.

\section{Parameterized PDEs}

Let $u: \tSpace \times \xSpace \times \mSpace  \to \R$ be a solution field that represents, e.g., temperature, density, velocities, or pressure of a physical process. The solution field $u$ depends on time $t \in \tSpace = [0,\Tend) \subset \R$, spatial coordinate $\bfx \in \xSpace \subset \R^\dimX$, and physics parameter $\m \in \mSpace \subset \mathbb{R}^{d^{\prime}}$. %
The solution field $u$ is governed by a parameterized PDE,
\begin{equation}
  \label{eq:pde}
  \begin{aligned}
    \partial_t u(\t, \x; \m)= & f(\x, u; \m) & \text { for }(\t, \x) \in  \tSpace  \times \xSpace \\
    u(0, \x; \m)=             & u_0(\x; \m)  & \text { for } \x \in \xSpace                       \\
  \end{aligned}
\end{equation}
where $u_0$ is the initial condition and $f$ can include partial derivatives of $u$ in $\bfx$. In the following, we always have appropriate boundary conditions so that the PDE problem \eqref{eq:pde} is well posed. The physics parameter $\m$ can enter the dynamics via $f$ and the initial condition $u_0$. Standard numerical methods such as finite-element \cite{Hughes2012} and finite-volume \cite{LeVeque_2002} methods can be used to numerically solve \eqref{eq:pde} to obtain a numerical solution $\fom(\cdot, \cdot; \m): \tSpace \times \xSpace \to \mathbb{R}$ for a physics parameter $\m \in \mSpace$.

Computational procedures for learning reduced models \cite{antoulas2005approximation,RozzaPateraSurvey,benner2015survey,interpbook,P22AMS,doi:10.1146/annurev-fluid-121021-025220} are typically split into an offline and an online phase: In the offline (training) phase, the reduced model is constructed from training trajectories
\begin{equation}\label{eq:FOMData}
  \fom(\cdot, \cdot; \bfmu_1), \dots, \fom(\cdot, \cdot; \bfmu_{\nm}): \xSpace \times \tSpace \to \R
\end{equation}
over offline physics parameters $\bfmu_1, \dots, \bfmu_{\nm} \in \mSpace$, which have been computed with the high-fidelity numerical model. In the subsequent online phase, the reduced model is used to rapidly predict solution fields at new physics parameters and initial conditions. %

\section{CoLoRA neural networks}
We introduce CoLoRA networks that (a) provide nonlinear parameterizations that circumvent the Kolmogorov barrier of linear model reduction and (b) impose an inductive bias that treats the time variable $\t$ differently from the spatial variable $\bfx$, which reflects that time is a special variable in physics. In particular, CoLoRA networks allow a continuous adaptation of a low number of network weights over time to capture the dynamics of solution fields (``fine-tuning'') for different physics parameters. %
\subsection{LoRA layers}

CoLoRA networks are motivated by LoRA \cite{hu_lora_2021}, a method that has been introduced to fine-tune large language models on discrete downstream tasks. LoRA layers are defined as
\begin{equation}
  \label{eq:coloralayer}
  \mathcal{C}(\bfx) = \W \bfx + \Delta\W\bfx + \bfb
\end{equation}
where $\bfx \in \R^{\dimX}$ is the input vector, $\W, \Delta\W \in \mathbb{R}^{n \times \dimX}$ are weight matrices, and $\bfb \in \mathbb{R}^{n}$ is the bias term.
The key of LoRA is that only $\Delta\W$ is changed during fine tuning and that $\Delta\W$ is of low rank $\rank \ll \min\{n,\dimX\}$ so it can be parameterized as,
\[
  \Delta\W = \A\B\,,\qquad \A \in \mathbb{R}^{n \times \rank}, \B \in \mathbb{R}^{\rank \times \dimX}\,.
\]
Thus, only $n \times \rank + \rank \times \dimX \ll n \times \dimX$ parameters need to be update per layer during fine-tuning rather than all $n \times \dimX + n$ as during pre-training,

\subsection{The CoLoRA layer}
Models with low intrinsic dimension are very common not only in large language models but also in many applications in science and engineering with phenomena that are described by PDEs \cite{10.1093/imanum/drx052,https://doi.org/10.1002/gamm.201310004,berman_randomized_2023}.
However, in the PDE settings, we have the special time variable $\t$ that requires us to ``fine-tune'' continuously as the PDE solution trajectories evolve; see Figure~\ref{fig:CoLoRAOverview}. Additionally, time imposes causality, which we want to preserve in CoLoRA models.

To enable a continuous low-rank adaptation, we introduce Continuous LoRA (CoLoRA) layers,
\begin{equation}
  \colora(\bfx) = \W \bfx + \alpha(\t; \m) \A\B \bfx + \bfb   \label{eq:coloralayer_small}
\end{equation}
where $\Delta\W = \A\B$ is a low-rank matrix of rank $\rank$ that is trained offline and $\alpha(\t; \m) \in \R$ is the online (``fine-tuning'') parameter that can change continuously with $\t$ and also with the physics parameter $\m$ in the online phase of model reduction. For example, when using a multilayer perceptron (MLP) with $\clayers$-many $\colora$ layers and a linear output layer $\boldsymbol{c} \in \mathbb{R}^{n}$, we obtain
\begin{equation}\label{eq:CoLoRAMLP}
  \hat{u}(\bfx; \paramsOff, \paramsOn(t, \m)) = \boldsymbol{c}^T(\colora_1(\sigma(\colora_2(\dots \sigma(\colora_L(\bfx))\dots))
\end{equation} with activation function $\sigma$ and $\colora_i(\bfx) = \W_i\bfx + \alpha_i(t, \m)\A_i\B_i\bfx + \bfb_i$ for $i = 1, \dots, \clayers$. The online parameters are given by the vector $\paramsOn(t, \m) = [\alpha_1(t, \m), \dots, \alpha_{\dimReduced}(t, \m)] \in \mathbb{R}^{\dimReduced}$ with $\dimReduced = \clayers$ in the example \eqref{eq:CoLoRAMLP}. We will later refer to $\paramsOn(t, \mu)$ as the latent state. All other CoLoRA parameters that are independent of time $\t$ and physics parameter $\m$ are trainable offline and collected into the offline parameter vector $\paramsOff \in \mathbb{R}^{\dimParamsOff}$ of dimension $\dimParamsOff \gg \dimReduced$.

We note that in principle $\A\B$ could be full rank without increasing the size of $\dimReduced$. But this would increase the number of parameters in $\paramsOff$. Additionally, the authors of LoRA \cite{hu_lora_2021}  observed that full rank fine-tuning updates under-perform low rank ones despite having more degrees of freedom, which is also in agreement with the low ranks used in dynamic low-rank and online adaptive methods in model reduction \cite{doi:10.1137/050639703,SAPSIS20092347,Peherstorfer15aDEIM,doi:10.1137/18M1218686,P18AADEIM,uy2022reduced,singh_lookahead_2023}.

A CoLoRA network defines a function
$\hat{u}: \xSpace \times \R^{\dimParamsOff} \times \R^{\dimReduced} \to \mathbb{R}$
that depends on an input $\bfx \in \Omega$, which is the spatial coordinate in our PDE setting, the offline parameters $\paramsOff \in \mathbb{R}^{\dimParamsOff}$ that are independent of time $t$ and physics parameter $\bfmu$ and the online parameters or latent state $\paramsOn(\t, \m) \in \mathbb{R}^{\dimReduced}$ that depends on $\t$ and $\m$. A CoLoRA network can also output more than one quantity by modifying the output layer, which we will use for approximating systems of PDEs in the numerical experiments.
A CoLoRA network $\rom$ is an implicit neural representation \cite{sitzmann2020implicit, brunton_implcitflow} in the sense that the PDE solution field is given implicitly by the parameters $\paramsOff$ and $\paramsOn(t, \m)$ and it can be evaluated at any coordinate $\bfx \in \xSpace$, irrespective of discretizations and resolutions used during training. %

If $\alpha_i(t; \m)$ is a scalar for each layer $i = 1, \dots, \clayers$, then the dimension of $\paramsOn(t, \m)$ equals the number of layers in $\rom$ in the MLP example \eqref{eq:CoLoRAMLP}, which can be overly restrictive. So we additional allow to have $\rank$-many online parameters $\alpha_1(t, \m), \dots, \alpha_{\rank}(t, \m)$ for each $\colora$ layer, in which case we have $
  \colora(\bfx) = \W \bfx + \A \diag({\alpha_1(\t; \m), \dots,  \alpha_{\rank}(\t; \m)}) \B \bfx + \bfb.$
The dimension of the online parameter vector $\paramsOn(t, \bfmu)$ is then $\dimReduced = \rank \times \clayers$. Other approaches are possible to make CoLoRA networks more expressive such as allowing $\A$ and $\B$ to depend on $\t$ and $\m$ as well; however, as we will show with our numerical experiments, only very few online parameters are needed in our experiments. %

Other works have examined similar weight matrix decomposition to separate out a set of adaptable parameters in the context of PDEs \cite{cho2023hypernetworkbased, wen_reduced-order_2023} and in implicit neural representations \cite{kim_generalizable_2023}; however, none directly treat the low rank adaptation as a function of time, which is key for our approach.

\subsection{CoLoRA networks can circumvent the Kolmogorov barrier}\label{sec:CoLoRANWidth}
CoLoRA networks can be nonlinear in the online parameter $\paramsOn(t, \m)$ and thus achieve faster error decays than given by (linear) Kolmogorov $n$-widths.
We give one example by considering the linear advection equation $\partial_t u(t, x) + \mu\partial_x u(t, x) = 0$ as in \citet{Ohlberger16} with initial condition $u(0, x) = u_0(x)$ and solution $u(x, t; \mu) = u_0(x - t\mu)$, which can lead to a slow $n$-width decay of $1/\sqrt{n}$ for a linear parameterization with $n$ parameters. In contrast, with the CoLoRA MLP network $\rom(x; \paramsOff, \paramsOn(t, \mu)) = c(\colora_1(\sigma_1(\colora_2(\sigma_2(\colora_3(x))))))$ with $\clayers = 3$ layers, we can exactly represent translation and thus the solution $u(t, x; \mu) = u_0(x - t\mu)$ of the linear advection equation: Set $\W_3 = [1, 0]^T, \bfb_3 = [0, 1]^T, \alpha_3(t, \m) = 0$ and $\A_3, \B_3$ arbitrary. Further $\W_2 = [1, 0], \bfb_2 = [0], \alpha_2(t, \m) = -t\m$ and $\A_2\B_2 = [0, 1]$ and $\W_1 = [1], \bfb_1 = [0], \alpha_1(t, \m) = 0$. If we use the known initial condition as activation function $\sigma_1 = u_0$ and the identity as activation function $\sigma_2$ and set $c = 1$, then we obtain $\hat{u}(x, \paramsOff,\paramsOn(t, \m)) = u_0(x - t\mu)$, which is the solution of the linear advection example above. Note that using the initial condition as an activation function is proper in this context because the initial condition is typically given in closed form or at least can be evaluated over $x$ and thus can be fitted during the pre-training. 

Of course this example with the linear advection equation is contrived but it shows how translation can be represented well by CoLoRA networks, which is the challenge that leads to the Kolmogorov barrier \cite{P22AMS}.
A more detailed treatment of the approximation theoretic properties of CoLoRA networks remains an open theory question that we leave for future work.

\section{Training CoLoRA models offline}

The goal of the following training procedure is to learn the offline parameters $\paramsOff$ of a CoLoRA network $\rom$ for a given parameterized PDE \eqref{eq:pde} so that only the much lower dimensional latent state $\paramsOn(t, \bfmu)$ has to be updated online over time $t$ and physics parameters $\m$ to approximate well the solution of the PDE. %

\paragraph{Enforcing continuity in time}
CoLoRA models make a careful treatment of time $t$, which enters via the latent state $\paramsOn(t, \m)$ rather than as input as the spatial coordinates $\bfx$; see also the discussion in Section~\ref{sec:Intro}.
We want that special meaning of the time variable to be also reflected in the pre-training approach in the sense of imposing regularity in the latent state $\paramsOn(t, \m)$ with respect to time $t$. Having smooth latent dynamics is a key property that has many desirable outcomes such as rapid time-stepping with large time step sizes, stability, and robustness to numerical perturbations.
A naive pre-training of CoLoRA over $\paramsOff$ and $\paramsOn$ with a global optimization problem would allow $\paramsOn$ to change arbitrarily, i.e., in a non-smooth way, over time. %

To impose regularity with respect to $t$, we introduce a hyper-network $h: \tSpace \times \mSpace \times \mathbb{R}^{\dimHyper} \to \mathbb{R}^{q}$ that depends on the parameter vector $\paramsHyper \in \mathbb{R}^{\dimHyper}$; see \cite{belbute-peres2021hyperpinn,brunton_implcitflow,cho2023hypernetworkbased} for other methods that build on hyper-networks in different settings.
Time $t$ and physics parameter $\m$ are inputs to the hyper-network and $\paramsOn(t, \m)$ is the output.
We focus on the case where $h$ is an MLP, but we stress that any other regression model can be used that provides the necessary regularity from $t$ to $\paramsOn(t, \m)$. For our purposes here, it is sufficient to choose $h$ such that it is continuous in $t$. This means that its output written as $\paramsOn(\t; \m)$ will depend continuously on time $t$.

\paragraph{Loss function for pre-training}

Recall that we have access to training data in the form of solution fields for $\bfmu_1, \dots, \bfmu_{\nm}$ given in \eqref{eq:FOMData}. For $i = 1, \dots, \nm$, we consider finite sets $\xSamples_i \subset \Omega$ and $\tSamples_i \subset \tSpace$ of spatial coordinates $\bfx$ and time $t$ samples over the spatial domain and time domain, respectively. For each offline physics parameter $\m_i$, we consider the relative error over the cross product $\xSamples_i \times \tSamples_i$
\[%
  J_i(\paramsOff, \paramsHyper) = \sum_{\substack{\bfx \in \xSamples_i\\ t_i \in \tSamples_i}} \frac{|\fom(\bfx, t; \bfmu_i) - \rom(\bfx; \paramsOff, h(t, \bfmu; \paramsHyper))|^2}{|\fom(\bfx, t; \bfmu_i)|^2}\,,
\]
where $\fom$ denotes the training trajectory for $\m_i$ that is available from the high-fidelity numerical model, $\rom$ is our CoLoRA parameterizations, and $h$ is the hyper-network discussed in the previous paragraph.

The loss that we optimize for the offline parameters $\paramsOff \in \mathbb{R}^{\dimParamsOff}$ and the parameter vector $\paramsHyper \in \mathbb{R}^{\dimHyper}$ of the hyper-network $h$ averages the relative errors $J_i$ over all training physics parameters,
\begin{equation}
  \label{eq:loss}
  L(\paramsOff, \paramsHyper) = \frac{1}{\nm}\sum\nolimits_{i = 1}^{\nm} J_i(\paramsOff, \paramsHyper)\,.
\end{equation}
This is also the mean relative error that we report on test parameters in our experiments; see Section~\ref{sec:numerics}.

We stress that in the spirit of implicit neural representations, the pre-training (as well as online) approach is independent of grids in the spatial and time domain. In fact, our formulation via the sets $\xSamples_i$ and $\tSamples_i$ for each training physics parameter $i = 1, \dots, \nm$ allows for different, unstructured samples in the spatial and time domain for each training physics parameter.

\begin{figure*}[t!]
\centering
\input{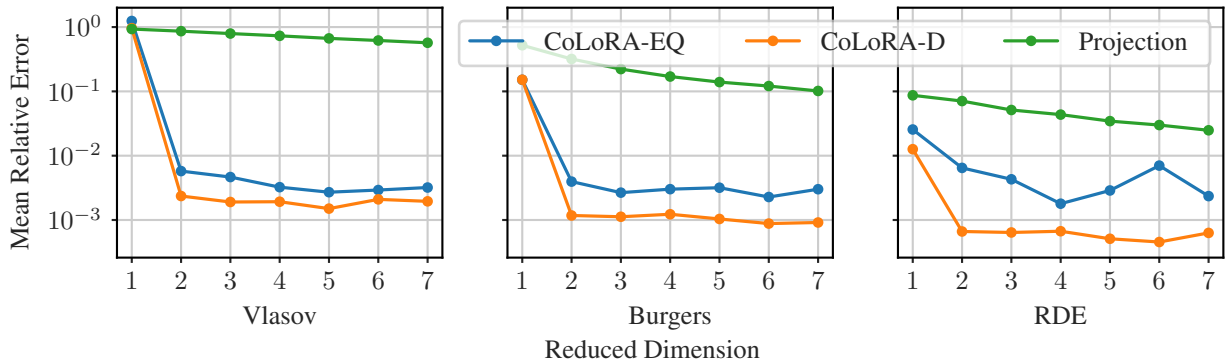}
\caption{CoLoRA models achieve orders of magnitude lower errors than the best-approximation error of linear model reduction methods, which is in agreement with Section~\ref{sec:CoLoRANWidth} that states that CoLoRA parameterizations circumvent the Kolmogorov barrier.}
\label{fig:nwidth}
\end{figure*}

\section{Online phase of CoLoRA models}
\label{sec:onlinecolora}
Given a new parameter $\m$ that we have not seen during training, the goal of the online phase is to rapidly approximate the high-fidelity numerical solution $\fom$ at $\m$. With pre-trained CoLoRA models, we can go about this in two fundamentally different ways: First, we can take a purely data-driven route and simply evaluate the hyper-network $h$ at the new $\m$ at any time $t$.  Second, because the latent state $\paramsOn(t, \m)$ depends on time $t$, we can take an equation-driven route and use the governing equation given in \eqref{eq:pde} to derive the online parameter $\paramsOn(\t, \m)$ via a variational formulation such as Neural Galerkin schemes \cite{bruna_neural_2024}; see \cite{lasser_lubich_2020,Du_2021,doi:10.1137/21M1415972,berman_randomized_2023} for other sequential-in-time methods that could be combined with CoLoRA.

\subsection{Data-driven forecasting (CoLoRA-D)}
We refer to CoLoRA models as CoLoRA-D if predictions at a new physics parameter $\m$ are obtained by evaluating the hyper-network $h$ at $\m$ and the times $\t$ of interest. The predictions that are obtained from CoLoRA-D models are purely data-driven and therefore do not directly use the governing equations in any way; neither during the pre-training nor during the online phase. %
Reduced models based on CoLoRA-D are non-intrusive \cite{Ghattas_Willcox_2021,doi:10.1146/annurev-fluid-121021-025220}, which can have major advantages in terms of implementation and deployment because only data needs to be available; these advantages are the same as for operator learning \cite{li_fourier_2021,Lu2021} that also is non-intrusive and typically relies only on data rather than the governing equations. The accuracy of CoLoRA-D models, however, critically depends on the generalization of $h$, which is in agreement with data-driven forecasting in general that has to rely on the generalization of data-fitted functions alone.

\subsection{Equation-driven predictions (CoLoRA-EQ)}
If the governing equations given in \eqref{eq:pde} are available, they can be used together with CoLoRA models to compute the states $\paramsOn(t, \bfmu)$ for a new parameter $\bfmu$ in a variational sense. We follow Neural Galerkin schemes \cite{bruna_neural_2024}, which provide a method for solving for parameters that enter non-linearly so that the corresponding parameterizations, in a variational sense, solve the given PDE.
We stress that such a variational formulation is possible with CoLoRA models because the latent state $\paramsOn(t, \m)$ depends on time $t$  rather than time being an input as the spatial coordinate $\bfx$. In particular, the sequential-in-time training of Neural Galerkin schemes is compatible with the time-dependent online parameter $\paramsOn$. Together with Neural Galerkin schemes, CoLoRA provides solutions that are causal, which is different from many purely data-driven methods. %

Neural Galerkin schemes build on the Dirac-Frenkel variational principle \cite{dirac_1930,Frenkel1934,lubich2008quantum,lasser_lubich_2020}, which can be interpreted as finding time derivatives $\dot{\paramsOn}(t, \m)$ that solve the Galerkin condition
\[
  \langle \partial_{\phi_i}\rom, r_t(\paramsOn(t, \m), \dot{\paramsOn}(t, \m)\rangle = 0\,, \quad i = 1, \dots, \dimReduced\,,
\]
so that the residual $r_t(\paramsOn, \dot{\paramsOn}) = \partial_t \rom - f(t, \cdot, \rom)$ of the PDE \eqref{eq:pde} as a function over the spatial domain $\Omega$ is orthogonal to the tangent space of the manifold $\{\rom(\cdot; \paramsOff, \paramsOn) \,|\, \paramsOn \in \mathbb{R}^{\dimReduced}\}$ induced by the online parameters at the current function $\rom(\cdot; \paramsOff, \paramsOn)$; we refer to \cite{bruna_neural_2024} for details and to Appendix~\ref{sec:NGComp} for the computational procedure.

The key feature of the equation-driven approach for predictions with CoLoRA models is that the latent states are optimal in a Galerkin sense, which provides a variational interpretation of the solution $\rom$ and opens the door to using residual-based error estimators to provide accuracy guarantees, besides other theory tools. Additionally, as mentioned above, it imposes causality, which is a fundamental principle in science that we often want to preserve in numerical simulations. Furthermore, using the governing equations is helpful to conserve quantities such as energy, mass, momentum, which we will demonstrate in Section~\ref{sec:NumExp:Physics}.

\begin{figure*}[t]
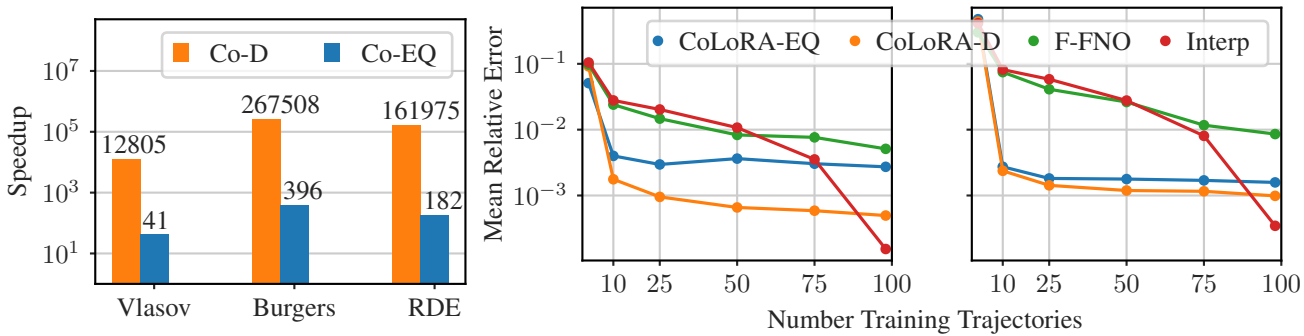

\centering
\begin{minipage}{0.3\textwidth}
\input{plots/bar_speedup_rel_1_3.pgf}
\end{minipage}
\hfill
\begin{minipage}{0.63\textwidth}
\input{plots/data_err_2_3.pgf}
\end{minipage}
\caption{\textbf{Left}: Purely data-driven CoLoRA (CoLoRA-D) is more than four orders of magnitude faster than traditional numerical models. If the governing equations are solved with Neural Galerkin \cite{bruna_neural_2024} in a Galerkin-optimal variational sense in the CoLoRA parameterization (CoLoRA-EQ), we still obtain about two orders of magnitude speedups while maintaining causality in the solution. \textbf{Right}: CoLoRA is more data efficient than operator learning and thus well suited for low-data regimes (Burgers', Vlasov).}
\label{fig:speed_data}
\end{figure*}

\section{Numerical experiments}
\label{sec:numerics}
\subsection{PDE problems}
\label{sec:equations}
The following three problems are challenging to reduce with conventional linear model reduction methods because the dynamics are transport dominated \cite{P22AMS}. %
Additional details on these equations and the full order models are provided in Appendix~\ref{appx:foms}.\\
\hspace*{0.2cm}\textbf{Collisionless charged particles in electric field}
The Vlasov equation describes the motion of collisionless charged particles under the influence of an electric field. %
We consider the setup of \citet{vlasovfom_2014}, which demonstrates filamentation of the distribution function of charged particles as they are affected by the electric field. Our physics parameter $\mu \in [0.2,0.4]$ enters via the initial condition. The full numerical model benchmarked in Figure~\ref{fig:speed_data} uses second-order finite differences on a $1024 \times 1024$ grid with adaptive time integrator. \\
\hspace*{0.2cm}\textbf{Burgers' equation in 2D}
Fields governed by the Burgers' equations can form sharp advecting fronts. The sharpness of these fronts are controlled by the viscosity parameter which we use as our physics parameter $\mu \in [10^{-3}, 10^{-2}]$. The full model benchmarked in Figure~\ref{fig:speed_data} uses 2nd-order finite differences and a $1024 \times 1024$ spatial grid with an implicit time integration scheme using a time step size of $\num{1e-3}$. \\
\hspace*{0.2cm}\textbf{Rotating denotation waves} We consider a model of rotating detonation waves, which is motivated by space propulsion with rotating detonation engines (RDE) \cite{rde_1, rde_2, rde_3}. 
The physics parameter $\mu$ 
we reduce over corresponds to the combustion injection rate, which leads to bifurcation phenomena that we investigate over the interval $\mu \in [2.0,3.1]$. The full model benchmarked in Figure~\ref{fig:speed_data} uses a finite volume method on a $2048$ grid with an implicit time integration scheme using a time step size of $\num{1e-3}$. \\
\hspace*{0.2cm}\textbf{Other PDE models} We also look at other PDEs to benchmark against methods from the literature; see Table~\ref{tbl:comparisons}. These include a two-dimensional wave problem with a four-dimensional physics parameter taken from \citet{yin_continuous_2023} and a three-dimensional shallow water wave example from \citet{serrano2023operator}. %

\subsection{CoLoRA architectures}
\label{sec:arch}
The reduced-model parameterization $\rom$ is an MLP with CoLoRA layers. The hyper-network $h$ is an MLP with regular linear layers. Both use swish activation functions. The most important architectural choice we make is the size of our networks—$\rom$ has 8 layers each of width 25 and $h$ has three layers each of width 15. As discussed earlier, such small networks are sufficient because of the strong inductive bias and low-rank continuity in time of CoLoRA networks. Only for the 3D shallow water example we use layer width 128. The larger width helps to capture the oscillations in the solution field in this example; see also Section~\ref{sec:Conc}. The error metric we report is the mean relative error, which is also our loss function \eqref{eq:loss}. More details are in Appendix~\ref{appx:architecture} and Figure~\ref{fig:tikzcolora}.

\subsection{CoLoRA and number of latent parameters}

Figure~\ref{fig:nwidth} compares the mean relative error of the proposed CoLoRA models with conventional linear projections, which serve as the empirical best-approximation error that can be achieved with any linear model reduction method (see Appendix~\ref{appx:LinearProj}). %
In all examples, the error is shown for test physics parameters that have not been used during training; see Appendix~\ref{appx:traintestdata}.

First, the linear approximations are ineffective for all three examples, which is in agreement with the observation from Section~\ref{sec:equations} that the three PDE models are challenging to reduce with linear model reduction methods. Second, our CoLoRA models achieve orders of magnitude lower relative errors for the same number of parameters as linear approximations. In all examples, 2--3 latent parameters are sufficient in CoLoRA models, which is in agreement with the low dimensionality of the physics parameters of these models. After the steep drop off of the error until around $\dimReduced = 2$ online parameters, there is a slow improvement if any as we increase $\dimReduced$, which is in agreement with other nonlinear approximations methods \cite{chen_crom_2023, carlburg_lee_2020}. This is because once $\dimReduced$ is equal to the intrinsic dimension of the problem, compression no longer helps reduce errors in predictions and instead the error is driven by other error sources such as time integration and generalization of the hyper-network. In these examples, the purely data-driven CoLoRA-D achieves slightly lower relative errors than the equation-driven CoLoRA-EQ, which could be due to the time integration error. In any case, the CoLoRA-EQ results show that we learn representations that are consistent with the PDE dynamics in the sense of Neural Galerkin based on the Dirac-Frenkel variational principle.

\subsection{Speedups of CoLoRA} %

In Figure~\ref{fig:speed_data}, we show the relative speedup of CoLoRA when compared to the runtime of the high-fidelity numerical models based on finite-difference and finite-volume methods as described in Appendix~$\ref{appx:foms}$. The speedups in the Burgers' and the RDE examples are higher than in the Vlasov example because we use an explicit time integration scheme for Vlasov but implicit ones for Burgers' and RDE. %
When integrating the governing equations in CoLoRA-EQ, we achieve speedups because of the smoothness of the latent dynamics of $\paramsOn(\t;\mu)$ as shown in Figure~\ref{fig:motivation}. The smoothness allows us to integrate $\paramsOn(\t;\mu)$ with a solver that uses an adaptive time-step control, which adaptively selects large time steps due to the smoothness of the dynamics. %
When using CoLoRA-D, we achieve orders of magnitude higher speedups because forecasting requires evaluating the hyper-network $h$ only. This can be done quickly due to the small size of the hyper-network $h$ as described in Section~\ref{sec:arch}. We note that we benchmark our method on the time it takes to compute the latent state $\paramsOn(\t;\mu)$ on the same time grid as the full model. %
There will of course be additional computational costs associated with plotting the CoLoRA solution on a grid in $\Omega$. %

\subsection{Data efficiency versus operator learning}
\label{sec:dataeff}

A key difference to operator learning is that CoLoRA aims to predict well the influence of the physics parameter $\m$ on the solution fields, rather than aiming for a generic operator that maps a solution at one time step to the next. We now show that CoLoRA can leverage the more restrictive problem formulation so that fewer training trajectories are sufficient. As Figure~\ref{fig:speed_data} shows for the Burgers' and Vlasov example, we achieve relative errors in the range of $\num{1e-03}$ with only about $\nm = 10$ training trajectories, whereas the operator- learning variant F-FNO \cite{tran_factorized_2023} based on Fourier neural operators (FNOs) \cite{li_fourier_2021} leads to an about one order of magnitude higher relative error. %
Neural operators struggle to achieve relative errors below \num{1e-2}, while CoLoRA achieves one order of magnitude lower relative errors with one order of magnitude fewer training trajectories. We also compare to simply linearly interpolating the function $\fom$ over space, time, and parameter and observe that in low data regimes CoLoRA achieves orders of magnitude more accurate predictions. In high data regimes, for sufficiently smooth problems, linear interpolation becomes  accurate as training physics parameters start to be closer and closer to test physics parameters; see Appendix~\ref{appx:traintestdata} for details. %

\begin{table*}[t]
\centering
\def\arraystretch{1.3}%
\begin{tabular} {l | ll|ll }
 \midrule
example:   &  \multicolumn{2}{c|}{\textbf{three-dim.~spherical shallow water}} &  \multicolumn{2}{c}{\textbf{two-dim.~wave}} \\ 
metric:           & MSE &  number parameters  &                           MSE &  number parameters                        \\  \midrule
MP-PDE            & 9.37e-5 & -                                  & 9.256e-7  & -                        \\  \hline
DeepONet          & 6.54e-3 & -                                 & 1.847e-2   & -                       \\  \hline
DINo              & 4.48e-5 & \num{2022912}                               & 9.495e-6   & \num{579776}                        \\  \hline
CORAL             & 3.44e-6 &  \num{1049344}                               & -          & -                         \\  \hline
\textbf{CoLoRA-D} & \textbf{3.19e-06}      & \textbf{\num{335744}}                     & \textbf{1.891e-07}  & \textbf{\num{7505}}                \\  \midrule
\end{tabular}
\caption{CoLoRA is more accurate than a range of other methods for forecasting PDEs and model reduction based on implicit neural representations while using significantly fewer parameters. MSE values and parameter counts are taken from \citet{yin_continuous_2023} for the 2D wave problem and from \citet{serrano2023operator} for 3D spherical shallow water. Parameter counts are estimated from the descriptions architecture depth and width in the original papers.}
\label{tbl:comparisons}
\end{table*}

\subsection{Leveraging physics knowledge in the online phase with CoLoRA-EQ}\label{sec:NumExp:Physics}
In the numerical experiments conducted so far, the purely data-driven CoLoRA-D outperforms CoLoRA-EQ in terms of error and speedup. However, using the physics equations online can be beneficial in other ways such as for causality and theoretical implications, especially for residual-based error estimators; see Section~\ref{sec:onlinecolora}. We now discuss another one here numerically, namely conserving quantities during time integration.
\begin{figure}
  \centering
  \input{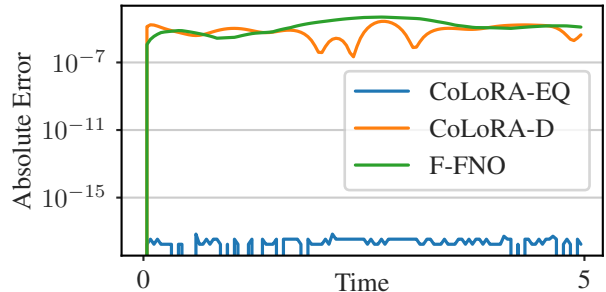}
  \caption{Solving the governing equations in a variational sense with Neural Galerkin \cite{bruna_neural_2024} and CoLoRA parameterizations (CoLoRA-EQ) leads to causal solutions and allows conserving quantities \cite{schwerdtner_nonlinear_2023} such as mass in the Vlasov problem, which is key for building trust in physics predictions and for interpretability.}
  \label{fig:conserve}
\end{figure}
We build on conserving Neural Galerkin schemes introduced in \citet{schwerdtner_nonlinear_2023} to conserve the mass of the probability distribution that describes the particles in the Vlasov problem. Preserving unit mass can be important for physics interpretations. In Figure~\ref{fig:conserve}, we show that using the CoLoRA-EQ with \citet{schwerdtner_nonlinear_2023}, we are able to conserve the mass of solution fields of the Vlasov equation to machine precision. By contrast, neither the CoLoRA-D nor F-FNO conserve the quantity, as the numerical results indicate. 

\subsection{Comparison to other nonlinear methods}

We run CoLoRA on two benchmark problems. The first is described in the DINo publication \cite{yin_continuous_2023}. This is a 2D wave problem where the four dimensional parameter $\mu$  affects the position and magnitude of the initial condition. The second is described in the CORAL publication \cite{serrano2023operator}. It is a shallow water equation formulated over a 3D spherical domain where the $\mu$ parameter nonlinearly affects the initial condition. We additionally report the accuracy of two other methods MP-PDE \cite{brandstetter2023message} and DeepONet \cite{Lu2021} both originally benchmarked in \citet{yin_continuous_2023} and \citet{serrano2023operator}. We report these results in Table~\ref{tbl:comparisons}. We see that on these two challenging benchmark problems CoLoRA achieves the lowest mean squared error. In terms of implementations, CoLoRA succeeds using a relatively simple modulation scheme and straightforward pre-training. 
CoLoRA also outperforms all other methods while using close to one to two orders of magnitude fewer parameters.

\section{Conclusions, limitations, and future work}\label{sec:Conc} CoLoRA leverages that PDE dynamics are typically continuous in time while evolving on low-dimensional manifolds. CoLoRA models provide nonlinear approximations and therefore are efficient in reducing transport-dominated problems that are affected by the Kolmogorov barrier. At the same time, CoLoRA is data efficient and requires only few training trajectories in our examples. The continuous-in-time adaptation of CoLoRA network weights leads to rapid predictions of solution fields of PDEs at new physics parameters, which outperforms current state-of-the-art methods. 

\emph{Limitations} First, the theoretical analysis for reduced models based on CoLoRA is currently limited. The preliminary results on overcoming the Kolmogorov barrier for a specific setup with the linear advection equation cannot be directly generalized to other problems and thus a more in-depth analysis is necessary. Second, there are applications where pre-training the CoLoRA network once and for all is insufficient, such as when predicting bifurcations that are not represented in the training data. Then, an online adaptive updating of the offline parameters is desired, for which efficient methods need to be developed. 

\emph{Future work} First, to well approximate solution fields with high-frequency oscillations, sharp gradients, and other non-smooth features, reduced modeling with CoLoRA can be combined with Fourier feature embeddings and periodic activation functions. 
Second, our hyper-network based method of modulation succeeds mainly when generalizing to examples which are in-distribution in terms of $\bfmu$ and $t$. Later work might seek to expand CoLoRA's method of parameter modulation to settings with neural ordinary differential equations and other methods that can enhance CoLoRA's extrapolation ability.
Third, a future direction is scaling reduced modeling with CoLoRA to higher-dimensional problems in both parameter and spatial domain. We expect that active data collection will be key for CoLoRA models to be efficient in high dimension. 

We provide an implementation of CoLoRA at \url{https://github.com/julesberman/CoLoRA}.


\section*{Acknowledgements}
The authors were partially supported by the National Science Foundation under Grant No.~2046521 and the Office of Naval Research under award N00014-22-1-2728. This work was also supported in part through the NYU IT High Performance Computing resources, services, and staff expertise.

\section*{Impact Statement} This paper presents work whose goal is to advance the field of Machine Learning. There are many potential societal consequences of our work, none which we feel must be specifically highlighted here.

\bibliography{refs.bib}
\bibliographystyle{icml2024}

\clearpage
\appendix
\onecolumn

\section{Literature review of nonlinear model reduction}\label{appx:MORLiterature}

There is a wide range of literature on model reduction; see \citet{antoulas2005approximation,RozzaPateraSurvey,benner2015survey,interpbook,doi:10.1146/annurev-fluid-121021-025220} for surveys and textbooks. We focus here on model reduction methods that build on nonlinear parameterizations to circumvent the Kolmogorov barrier \cite{P22AMS}. 

First, there is a range of methods that pre-compute a dictionary of basis functions and then subselect from the dictionary in the online phase \cite{doi:10.1080/13873954.2011.547670,Dihlmann2011,https://doi.org/10.1002/nme.4371,https://doi.org/10.1002/nme.3327,doi:10.1137/120873868,doi:10.1137/130924408,https://doi.org/10.1002/nme.4773,doi:10.1098/rsta.2021.0206}. However, once the dictionary has been pre-computed offline, it remains fixed and thus such dictionary-based localized model reduction methods are less flexible in this sense compared to the proposed CoLoRA approach. 

Second, there are nonlinear reduced modeling methods that build on nonlinear transformations to either recover linear low-rank structure that can be well approximated with linear parameterizations in subspace or that augment linear approximations with nonlinear correction terms. For example, the early work \citet{ROWLEY20001} shows how to shift bases to account for translations and other symmetries. Other analytic transformations are considered in, e.g., \citet{OHLBERGER2013901,doi:10.1137/17M1140571,ehrlacher19,QIAN2020132401,PAPAPICCO2022114687,BARNETT2022111348,https://doi.org/10.48550/arxiv.2205.02304,ISSAN2023111689}. The works by \citet{TaddeiShock,Cagniart2019} parameterize the transformation maps and train their parameters on snapshot data rather than using transformations that are analytically available. 

Third, there are online adaptive model reduction methods that adapt the basis representation during the online phase \cite{doi:10.1137/050639703,SAPSIS20092347,PhysRevE.89.022923,GERBEAU2014246,doi:10.1002/nme.4800,Peherstorfer15aDEIM,https://doi.org/10.1002/nme.4770,P18AADEIM,UngerTransformModes2020,doi:10.1137/16M1071493,RAMEZANIAN2021113882,HUANG2023112356}. An influential line of work is the one on dynamic low-rank approximations \cite{doi:10.1137/050639703,doi:10.1137/140967787,doi:10.1137/18M1218686,EINKEMMER2021110353,doi:10.1137/140967787,Musharbash:231216,MUSHARBASH2018135,hesthaven_pagliantini_rozza_2022} that adapt basis functions with low-rank additive updates over time and thus can be seen as a one-layer version of CoLoRA reduced models.

\section{Details on numerical experiments}
\label{appx:num}
In Figure~\ref{fig:motivation} we have three plots which show the benefits of the CoLoRA method.  In the left plot, we show each dimension of $\paramsOn$ as a function of time. These parameters were generated through time integration (CoLoRA-EQ). This shows even with integration we get smooth dynamics. In the middle plot, we traverse the latent space by generating samples in the two dimensional space spanned by the component functions of $\paramsOn(t; \m)$ and then evaluating $\rom$ at each of these points. In the right plot, we train CoLoRA on Vlasov with a reduced dimension of 2. We then show the magnitude of the PDE residual at grid of points in the two dimensional space spanned by the component functions of the learned $\paramsOn(t; \m)$. The magnitude of the PDE residual is given by the residual from solving the least squares problem given in \eqref{eq:ng-lstsq} at each of these grid points. When plotting the resulting field we see that CoLoRA learns a continuous region of low PDE residual along which the latent training trajectories lie. The inferred latent test trajectories lie in between training trajectories showing the generalization properties of CoLoRA which allows for an accurate time continuous representation of the solution.

\begin{figure}[t]
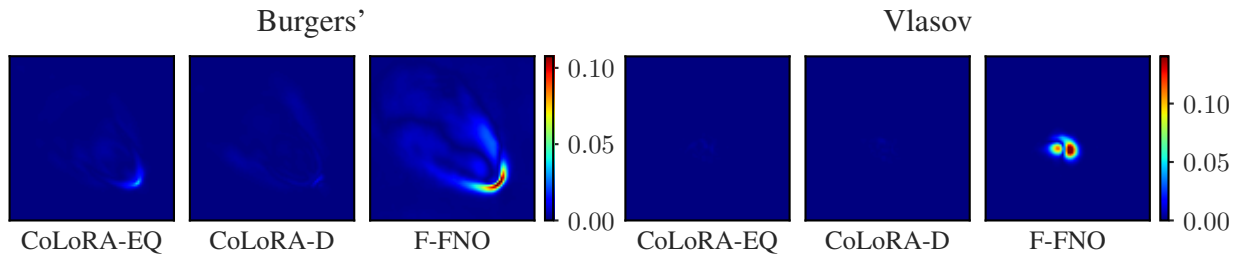

  \centering
  \import{plots}{errfield_burgers2d.pgf}
  \import{plots}{errfield_vlasovfix2.pgf}

  \caption{We show the point-wise absolute error of CoLoRA vs F-FNO. CoLoRA provides accurate solution fields even when trained on a low number of trajectories compared to operator learning. Plots here shown at 10 training trajectories.} %
  \label{fig:dataerrors}
\end{figure}

In Section~\ref{sec:dataeff} we report on the error of CoLoRA and F-FNO as a function of the number of training trajectories. In Figure~\ref{fig:dataerrors}, we give the point-wise error plots at 10 training trajectories. In particular we see that F-FNO has difficulty tracking the advection dynamics of the solution over time. CoLoRA by contrast is able to approximate these dynamics accurately.


All numerical experiments were implemented in Python with JAX \cite{jax2018github} with just-in-time compilation enabled. All benchmarks were run on a single NVIDIA RTX-8000 GPU.

\section{Description of full order models (FOMs)}

\label{appx:foms}
In order to ensure a fair comparison in terms of runtime between CoLoRA and the FOMs, we implement all FOMs in JAX \cite{jax2018github} with just-in-time compilation.

\subsection{Vlasov}
The Vlasov equations are
\[
  \partial_t u(\t,\bfx; \mu)=-x_2 \partial_{x_1} u(\t,\bfx; \mu)+  \partial_{x_1} \phi(x_1; \mu) \partial_{x_2} u(\t,\bfx; \mu)
\]
where $\bfx = [x_1, x_2]^T \in \R^2$. The first coordinate $x_1$ corresponds to the position of the particles and $x_2$ to the velocity. The potential of the electric field is $\phi(x)=-(0.2 + 0.2\cos(\pi x^4) + 0.1\sin(\pi x))$. We impose periodic boundary conditions on $\mathcal{X} = [-1, 1)^2$ and solve over the time domain $\tSpace = [0, 5]$. %
Our physics parameter $\mu \in [0.2,0.4]$ enters via the initial condition $u_0(\bfx;\mu) =  \exp(-100|(\x-0.2+\mu)|^2)\,.$

The Vlasov full order model uses a 4th order central difference stencil to compute spatial derivatives over a $1024 \times 1024$ spatial grid. This is then integrated using 5th order explicit Runge-Kutta method with an embedded 4th order method for adaptive step sizing.

\subsection{Burgers'}
The two-dimensional Burgers' equations are described as,
\begin{align*}
  \frac{\partial u}{\partial t} = -u \frac{\partial u}{\partial x}-v \frac{\partial u}{\partial y}+\mu \left(\frac{\partial^2 u}{\partial x^2}+\frac{\partial^2 u}{\partial y^2}\right) \\
  \frac{\partial v}{\partial t} = -u \frac{\partial v}{\partial x}-v \frac{\partial v}{\partial y}+\mu \left(\frac{\partial^2 v}{\partial x^2}+\frac{\partial^2 v}{\partial y^2}\right).
\end{align*}
We consider the spatial domain $ \mathcal{X} = [0, 1)^2$, time domain $\tSpace = (0, 1]$ where $\bfx = [x, y]^T \in \R^2$ and the physics parameter $\mu \in [10^{-3}, 10^{-2}]$ corresponds to the viscosity. We impose periodic boundary conditions with the initial condition $u_0(x) = v_0(x) = \exp(-(14\pi)^2(x-\pi/10)^4)$. We note that when $u_0(x) = v_0(x)$ the two variables will be equal for all time, so we can effectively consider this as a single variable problem over a two-dimensional spatial domain.

For the Burgers' full order model we follow the full order model described \cite{wang_2d_nodate}. This uses finite differences to compute the spatial derivatives and uses a fixed-time step implicit method with Newton iterations for time integration. For the full order model benchmark we choose a $1024 \times 1024$ spatial grid.

\subsection{Rotating Detonating Engine}\label{sec:Appx:RDE}
The equations for the RDE setup we investigate are given as follows:
\begin{align*}
  \frac{\partial}{\partial t} \eta(x, t)=    & -\eta(x, t) \frac{\partial}{\partial x} \eta(x, t)+v \frac{\partial^2}{\partial x^2} \eta(x, t) \\
                                             & +(1-\lambda(x, t)) \omega(\eta(x, t))+\xi(\eta(x, t)),                                          \\
  \frac{\partial}{\partial t} \lambda(x, t)= & \nu \frac{\partial^2}{\partial x^2} \lambda(x, t)+(1-\lambda(x, t)) \omega(\eta(x, t))          \\
                                             & -\beta(\eta(x, t) ; \mu) \lambda(x, t)\,.
\end{align*}
The function $\omega$ which models the heat release of the system is given by,
\[
  \omega(\eta(x, t))=k_{\mathrm{pre}} e^{\frac{\eta(x, t)-\eta_c}{\alpha}}.
\]
The function $\beta$ describes the injection term and is given by,
\[
  \beta(\eta(x, t) ; \mu)=\frac{\mu}{1+e^{r\left(\eta(x, t)-\eta_p\right)}}.
\]
$\xi(\eta(x, t))=-\varepsilon \eta(x, t)$ corresponds to the energy loss of the system. We examine these equations on a circular domain $ \xSpace = [0, 2\pi)$ over time $\tSpace = [0, 20]$.
The hyperparamters for these equations are given as follows:
$\nu=0.025$, $k_{\mathrm{pre}}=1$, $\alpha=0.3$, $\eta_c=1.1$, $\eta_p=0.5$, $r=5$, $\epsilon=0.11$. The initial condition is given by,
\begin{align*}
  \eta(x, 0)   & = 0.4 \exp(-2.25(x-\pi)^2)+1.0\, \\
  \lambda(x,0) & = 0.75\,.
\end{align*}

The implementation for the RDE full order model follows \cite{singh_lookahead_2023} which uses finite differences to compute the spatial derivatives and use a fixed-time step implicit method with Newton iterations for time integration.

\section{Pre-training and architecture details}\label{appx:architecture}
As stated in Section~\ref{sec:arch}, the reduced-model parameterization $\rom$ is a multilayer perceptron with CoLoRA layers. There are 8 layers with swish nonlinear activation functions in between each layer. The first layer is a periodic embedding layer as described in Appendix~\ref{appx:periodiclayer} which ensures the network obeys the periodic boundary condition of the PDEs we consider. This leaves the 7 subsequent layers available to be either CoLoRA layers or standard linear layers. If the dimension of the online parameters are less than 7 ($\dimReduced < 7$), then the CoLoRA layers are the first $\dimReduced$ most inner layers in order to increase their nonlinear effect. For all $\colora$ layers, the rank is $\rank = 3$, unless otherwise stated.

The width of all layers is 25 except the last whose width must be 1 in order to output a scalar field. The only larger network is used in the 3D spherical shallow water example where the width is 128. In the case of the RDE example and the 2D Wave example given in \citet{yin_continuous_2023} the last layer is of width 2 in order to output a field for each variable in the equation. The hyper-network $h$ is a multilayer perceptron of depth 3 which also uses swish nonlinear activation function. The width of each layer is 15, except the last layer whose width is $\dimReduced$, the dimension of the online parameter vector $\paramsOn(t, \m)$.

\begin{figure*}[t]
  \centering
  \includegraphics[width=1.0\textwidth]{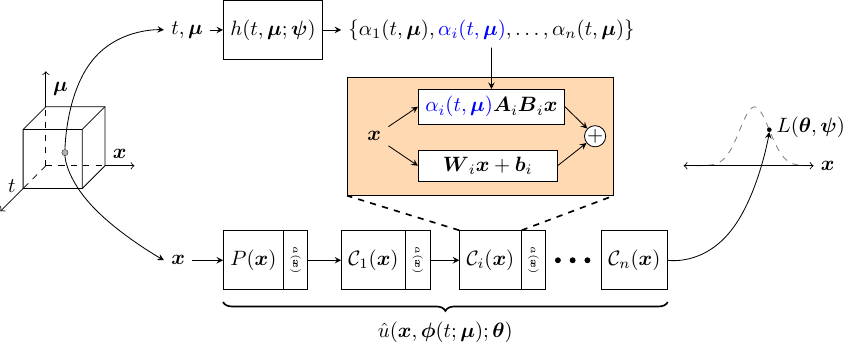}
  \caption{The CoLoRA architecture uses a hyper-network $h$ to generate a set of continuous parameters $\alpha$ which are used to scale low rank matrices $\A_i \B_i$ which are internal to the reduced order model $\rom$. The parameters of $\paramsHyper$ and $\paramsOff$ are then jointly optimized to fit data from the full order model $\fom$. }
  \label{fig:tikzcolora}
\end{figure*}

\subsection{Periodic $P$ layer}\label{appx:periodiclayer}
All the equations we consider here have periodic boundary conditions. These can be enforced exactly by having the first layer of $\rom$, which we call $P$, embed the $\x$ coordinates periodically. For an input $\x \in \R^\dimReduced$ a $P$ layer with period $\omega$ is defined as
$$
  P(\x)=\sum_{i=1}^d \left[ a\cos(\x\frac{2\pi}{\omega}+c)+b\right]_i
$$
where $a,c,b \in \R^d$ are additionally part of the offline parameters $\paramsOff$. The only exception is in 1D Inviscid Burgers' given in \cite{chen_crom_2023} which does not have periodic boundary conditions. Here we simply replace $P$ with another $\colora$ layer. In this case the boundary are loosely enforced via pretraining.
\subsection{Normalization}
The hyper-network given by $h$ normalizes its input so that $\m$ and $\t$ are mean zero and standard deviation 1, where these statistics are computed across the training data. The reduced model given by $\rom$ normalizes the $\x$ coordinates so that they are fixed between [0, 1]. The period of the periodic layer is then set to 1 in order to correspond to the normalized data.
\subsection{Pre-training}
In pre-training for all our benchmark problems (Vlasov, Burgers', and RDE) we minimize \eqref{eq:loss} using an Adam optimizer \cite{kingma2017adam} with the following hyper-parameters,
\begin{itemize}[nosep]
    \item learning rate :  $\num{5e-3}$
    \item scheduler : cosine decay
    \item $\beta_1$: 0.9
    \item $\beta_2$: 0.999
\end{itemize}
For the results given in Table~\ref{tbl:comparisons} for the 2D Wave and 3D Shallow Water problems, we use $\num{250000}$ and $\num{2000000}$ iterations respectively, with all other hyper-parameters kept the same.

\section{F-FNO experiments}
\label{appx:ffnoexp}
For implementation of the F-FNO we use the code base given in the original paper \cite{tran_factorized_2023} while keeping the modification that we make minimal. 
We use their largest architecture which is 24 layers deep as this was shown to give the best possible performance on their benchmarks. This was obtained via a grid sweep of the number of layers and time step size for the F-FNO. Additionally we give our $\m$ as input to their network. We train over 100 epochs as in their implementation. In order to give the F-FNO the best possible performance, the error reported is from the best possible checkpoint over all the epochs. All other hyper-parameters we set according to their implementation.

We provide additional results of the experiments in Table~\ref{table:dataffno}.

\begin{table*}[]
\centering
\def\arraystretch{1.3}%
\begin{tabular}{|l|lll|}
\hline
\textbf{Method}                              & \textbf{Vlasov (rel. err.)} & \textbf{2D Burgers (rel. err.)} & \textbf{RDE (rel. err.)} \\ \hline
\textit{High Data (100 Trajectories)} &                             &                                 &                          \\ \hline
F-FNO                                        & 8.57e-3                     & 5.11e-3                         & 2.21e-3                  \\ \hline
CoLoRA-EQ                                    & 1.58e-3                     & 2.27e-3                         & 1.49e-3                  \\ \hline
CoLoRA-D                                     & 9.87e-4                     & 4.96e-4                         & 2.05e-4                  \\ \hline
\textit{Low Data (10 Trajectories)}          &                             &                                 &                          \\ \hline
F-FNO                                        & 7.48e-2                     & 2.40e-2                         & 5.69e-3                  \\ \hline
CoLoRA-EQ                                    & 2.73e-3                     & 3.99e-3                         & 1.79e-3                  \\ \hline
CoLoRA-D                                     & 2.37e-3                     & 1.76e-3                         & 4.47e-4                  \\ \hline
\end{tabular}
\caption{Detailed results of F-FNO data efficiency experiment}
\label{table:dataffno}
\end{table*}

\section{Linear projection as comparison with respect to best approximation error}
\label{appx:LinearProj}
In order to compute the optimal linear projection we assemble the training and test data into two snapshot matrices. We then compute the singular value decomposition of the training snapshot matrix and build a projection matrix from the top $n$ left singular values where $n$ is the reduced dimension. We then use this projection matrix to project the test data into the reduce space and then project back up into the full space using the transpose of the projection matrix. We then measure the relative error between the resulting project test data and the original test data. This value gives the optimal linear approximation error. For additional details see \cite{doi:10.1146/annurev-fluid-121021-025220}.

\section{Sampling train and test trajectories}
\label{appx:traintestdata}
Section~\ref{sec:dataeff} examines the performance of CoLoRA against an F-FNO and linear interpolation as one increases the number of training trajectories. In order to appropriately run this experiment we need a consistent way of sampling the training trajectories from the $\m \in \mSpace$ ranges we examine. We first
generate many trajectories from equidistant-spaced parameters in our range $\mSpace$. This is our total trajectory dataset. We then pick three test trajectories from this set which are equally spaced out. Then as we increase the number of training trajectories (i.e. the value on the x-axis of Figure~\ref{fig:speed_data}), we pick trajectories from our total trajectory dataset so as to maximize the minimum distance of any training trajectory from any test trajectory. This ensures that as we increase the number of training trajectories the difficulty of the problem (from an interpolation perspective) decreases. %

For Burgers' we generate 101 equidistant samples of $\m$ in the range $\mSpace=[0.01, 0.001]$. For Vlasov we generate 101 equidistant samples of $\m$ in the range $\mSpace=[0.2, 0.4]$. The test samples for Burgers' are $[0.00253,0.0055,0.00847]$.  The test samples for Vlasov are $[0.234, 0.3, 0.366]$.

For all other experiments the train-test splits are as follows:

\begin{tabular}{|l l l|}\hline

  \textbf{Equation} & \textbf{Train}                                                        & \textbf{Test}        \\
  \hline
  Vlasov            & $[0.2, 0.224, 0.274, 0.3, 0.326, 0.376, 0.4]$                         & $[0.25, 0.35]$       \\
  Burgers           & $[0.001, 0.00199, 0.00298, 0.00496, 0.00595, 0.00694, 0.00892, 0.01]$ & $[0.00397, 0.00793]$ \\
  RDE               & $[2.0, 2.1, 2.2, 2.4, 2.5, 2.6, 2.8, 2.9, 3.0, 3.1]$                  & $[2.3, 2.7]$         \\
  \hline
\end{tabular}

\section{Neural Galerkin computational procedure}\label{sec:NGComp}
At each time step, for samples $\bfx_1, \dots, \bfx_{\nx} \in \Omega$, the computational procedure of Neural Galerkin schemes forms the batch gradient matrix $\boldsymbol{J}(\paramsOn(t;\m)) \in \R ^ {\nx \times \dimReduced}$ with respect to the online parameters,
\begin{align*}
  \boldsymbol{J}(\paramsOn(t;\m)) =
  [\nabla_{\phi_1}\rom(\bfx_1; \paramsOff,\paramsOn(t;\m)), \dots, \nabla_{\phi_{\dimReduced}}\rom(\bfx_{\nm}; \paramsOff, \paramsOn(t;\m))]^T
\end{align*}
and the $\nx$-dimensional vector $\boldsymbol{f}(t, \paramsOn(t, \m)) = [f(t, \bfx_1; \rom(\cdot; \paramsOff, \paramsOn(t, \m)), \dots, f(t, \bfx_{\nm}; \rom(\cdot; \paramsOff, \paramsOn(t, \m)))]$. The batch gradient and right-hand side lead to the linear least-squares problem in $\dot{\paramsOn}(\t, \m)$,
\begin{equation}
  \label{eq:ng-lstsq}
  \min_{\dot{\paramsOn}(\t; \m)}  \|  \boldsymbol{J}(\paramsOff, \paramsOn(t;\m)) \dot{\paramsOn}(\t; \m) - \boldsymbol{f}(\paramsOff, \paramsOn(t;\m)) \|^2_2\,,
\end{equation}
which is then discretized in time  and solved for the corresponding trajectory of latent states $\paramsOn(t_1, \m), \dots, \paramsOn(t_{\nt}, \m) \in \mathbb{R}^{\dimReduced}$ at the time steps $t_1 < \dots < t_{\nt}$. We refer to \cite{bruna_neural_2024,berman_randomized_2023} for details on this computational approach.

\end{document}